\pdfoutput=1

\documentclass[11pt]{article}

\usepackage[final]{acl}

\usepackage{times}
\usepackage{latexsym}

\usepackage[T1]{fontenc}

\usepackage[utf8]{inputenc}

\usepackage{microtype}

\usepackage{inconsolata}

\usepackage{amsmath}
\usepackage{amssymb}
\usepackage{mathtools}
\usepackage{amsthm}
\usepackage{algorithm}
\usepackage{algorithmic}
\usepackage{kotex}
\usepackage{multirow}
\usepackage{makecell}
\usepackage{booktabs}
\usepackage{graphicx}
\usepackage{adjustbox}
\usepackage{float}
\usepackage{wrapfig}
\usepackage{listings}
\usepackage{dblfloatfix}
\usepackage{tabularx}

\usepackage{longtable}
\usepackage{amssymb}
\usepackage{pifont}
\usepackage{xcolor}
\usepackage{dsfont}
\newcommand{\oursol}{\textnormal{SemGro}}

\usepackage{titlesec}
\usepackage{ragged2e}

\titleformat{\subsubsection}[runin]
{\normalfont\normalfont\bfseries}{\thesubsection}{1em}{}

\title{Semantic Skill Grounding for Embodied Instruction-Following in Cross-Domain Environments}

\author{Sangwoo Shin\textsuperscript{\rm 1}\thanks{Equally contributed to this work},
Seunghyun Kim\textsuperscript{\rm 1}\footnotemark[1], 
Youngsoo Jang$^2$, 
Moontae Lee$^{2, 3}$, 
Honguk Woo\textsuperscript{\rm 1}\thanks{Honguk Woo is the corresponding author
}
\\
    Department of Computer Science and Engineering, Sungkyunkwan University\textsuperscript{1} \\
    LG AI Research\textsuperscript{2} \hspace{0.4cm} University of Illinois Chicago\textsuperscript{3} \\
    \texttt{jsw7460@skku.edu, kimsh571@skku.edu} \\
    \texttt{youngsoo.jang@lgresearch.ai, moontae.lee@lgresearch.ai, hwoo@skku.edu}
  }

\lstdefinestyle{mypy}{
    language=Python,
    basicstyle=\ttfamily\small,
    showstringspaces=false,
    breaklines=true,
    frame=tb, 
    framerule=0.5pt,
    rulecolor=\color{black},
    numbers=none, 
    aboveskip=1em, 
    belowskip=1em, 
    columns=flexible,
    captionpos=t 
}

\begin{document}
\maketitle
\begin{abstract}
In embodied instruction-following (EIF), the integration of pretrained language models (LMs) as task planners emerges as a significant branch, where tasks are planned at the skill level by prompting LMs with pretrained skills and user instructions. However, grounding these pretrained skills in different domains remains challenging due to their intricate entanglement with the domain-specific knowledge. 
To address this challenge, we present a semantic skill grounding ($\oursol$) framework that leverages the hierarchical nature of semantic skills. $\oursol$ recognizes the broad spectrum of these skills, ranging from short-horizon low-semantic skills that are universally applicable across domains to long-horizon rich-semantic skills that are highly specialized and tailored for particular domains. 
The framework employs an iterative skill decomposition approach, starting from the higher levels of semantic skill hierarchy and then moving downwards, so as to ground each planned skill to an executable level within the target domain. To do so, we use the reasoning capabilities of LMs for composing and decomposing semantic skills, as well as their multi-modal extension for assessing the skill feasibility in the target domain. 
Our experiments in the VirtualHome benchmark show the efficacy of $\oursol$ in 300 cross-domain EIF scenarios.
\end{abstract}

\section{Introduction}
In the area of embodied instruction-following (EIF), the quest to develop systems capable of understanding user instructions and achieving tasks in a physical environment has been a pivotal aspiration for general artificial intelligence.
Recent advancements in language models (LMs) have demonstrated their extensive and impressive application in EIF, where LMs are employed as task planners, drawing on their commonsense knowledge of the physical world
~\cite{LACMA, padmakumar2023multimodal, EMMA, yun2023emergence, logeswaran2022few}.

The process of LM-based task planning in EIF unfolds via a two-stage pipeline: (i) Interpretable skills (i.e., semantic skills) are acquired, and (ii) given a user instruction, the LM is prompted with the instruction and the skill semantics to identify appropriate skills to perform. To ensure that such planned skill can be executed within the domain context the agent is placed in, several works have focused on incorporating environmental observations into task planning~\cite{progprompt, llmplanner, e2wm, gplanet, swiftsage, reflexion, glam}. Yet, EIF in cross-domain environments has not been fully explored. In cross-domain settings, the complexity goes beyond logical task planning by LMs; it necessitates the adaptation of pretrained skills to align with the specific characteristics and limitations of the target physical environment. This adaptation challenge stems from the inherent domain specificity of semantic skill acquisition, calling for a strategy that disentangles these learned behaviors from their original domain contexts and reapplies them in new domains.

In response to this challenge, we present a semantic skill grounding framework $\oursol$. We observe a trade-off between semantic depth and domain-agnostic executability within the skill hierarchy. High-level skills, such as ``prepare a meal'', embody a richer semantic content and long-horizon behavior compared to low-level skills, such as ``grab a cup''. However, these high-level skills tend to be effective only within specific domains, as their execution heavily depends on a multitude of environmental factors, unlike their low-level counterparts which can be universally applicable across domains. $\oursol$ capitalizes on this insight by employing an iterative skill grounding scheme, designed to progressively decompose pretrained skills to fit in the cross-domain EIF environment as well as to fully utilize the reasoning strengths of LMs.

To be specific, the iterative mechanism of $\oursol$ involves two procedures: (i) task planning with skills that align with a given instruction (in Section~\ref{subsec:taskplanner}) and (ii) assessing the executability of planned skills (in Section~\ref{subsec: critic}). 
In $\oursol$, a task planner employs in-context learning of LMs to interpret a given instruction and generate a series of semantic skills for the instruction through iterative skill grounding and decomposition. Concurrently, a skill critic employs the multi-modal extension of LMs to evaluate the applicability and execution potential of these skills within target domains. 
When the critic deems a planned skill as unsuitable for execution, the planner responds by generating an alternative, detailed instruction. This instruction is intended to be carried out through a sequence of lower-level skills that align with the previously planned skill. Through the collaborative interaction between the task planner and the skill critic, $\oursol$ navigates the semantic hierarchy of skills to establish the best trade-off between the logical relevance of skills to a given instruction and the skill feasibility in cross-domain environments.

%
%

Using the VirtualHome benchmark~\cite{virtualhome} to evaluate EIF agents~\cite{progprompt, saycanpay, llmstate, e2wm}, we demonstrate the advantages of $\oursol$ for cross-domain EIF scenarios. For instance, $\oursol$ outperforms the state-of-the-art task planning framework LLM-Planner~\cite{llmplanner} by $23.97\%$ in task completion performance.

Our contributions are three-fold:
\textbf{(i)} We present a novel framework $\oursol$ to address the issue of cross-domain skill grounding in EIF.
\textbf{(ii)} We devise an iterative skill grounding algorithm, leveraging the reasoning capabilities of LMs and their multi-modal extension.
\textbf{(iii)} We conduct an intensive evaluation of $\oursol$ in 300 cross-domain scenarios of the VirtualHome benchmark, complemented by ablation studies to explore the efficacy of $\oursol$.

\begin{figure}[t]
    \centering
    \includegraphics[width=1.0\columnwidth]{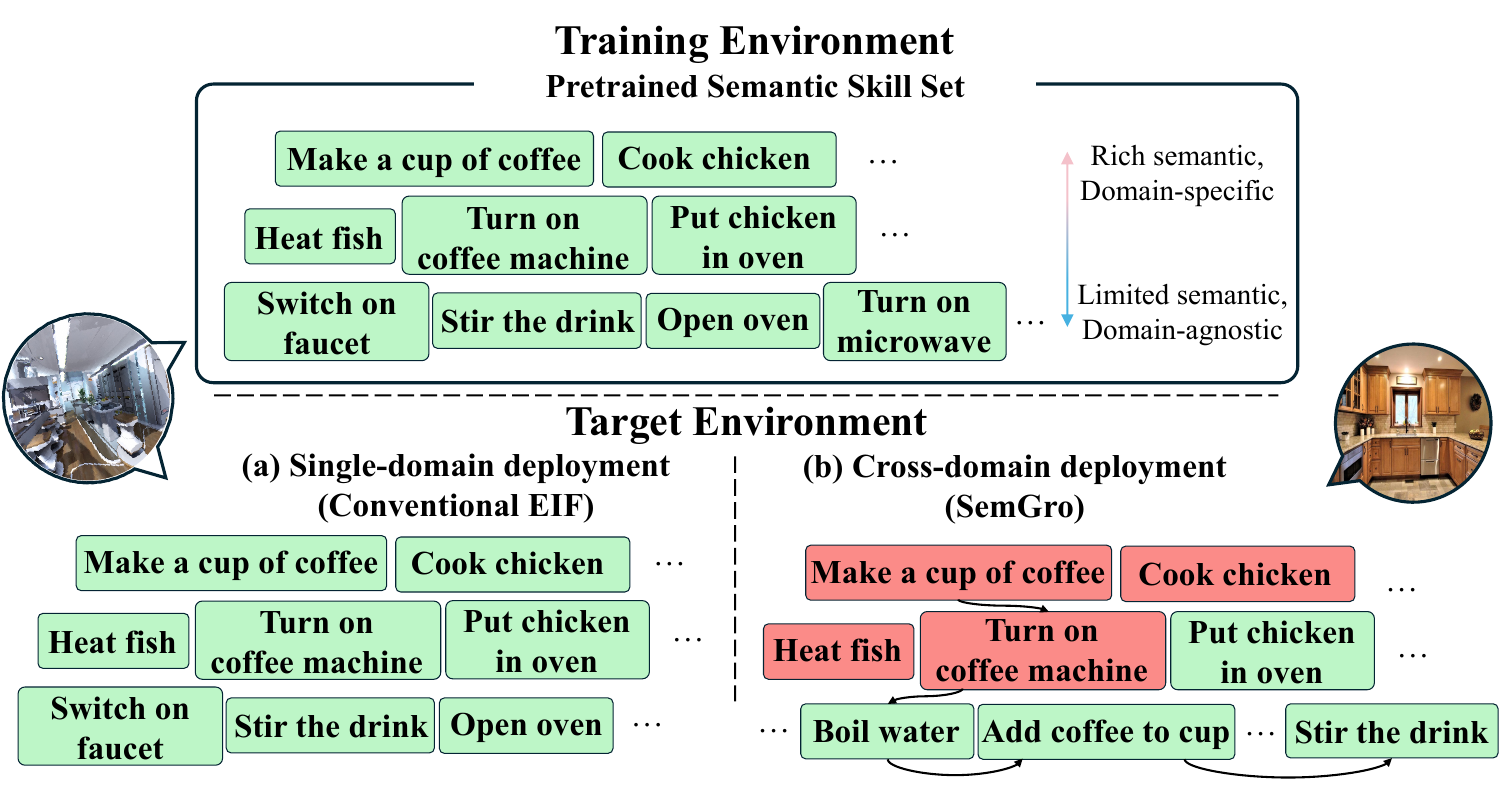}
    \caption{
    Cross-domain EIF. (a) Conventional approaches focus on \textbf{single-domain} deployment, where the pretrained skills are effective within only that particular domain (marked in the green boxes). 
    (b) Our $\oursol$ caters to \textbf{cross-domain} deployment, where the pretrained skills may be inexecutable (marked by red boxes) in the target environment. For this, $\oursol$ explores the semantic hierarchy of skills.
    }
    \label{fig:fig1}
\end{figure}

\section{Preliminaries}
\subsection{Language Models for Task Planning}
For complex embodied tasks, prior works have leveraged the notion of semantic skills, which represent interpretable patterns of expert behavior. These skills have a wide spectrum, ranging from short-horizon basic low-level skills (e.g., ``stir the drink'') to long-horizon high-level skills (e.g., ``prepare a meal''). These diverse skills $\Pi_l = \{\pi_l \, | \, l: \textnormal{skill semantic}\}$ can be efficiently established within a training environment, through deterministic algorithms, reinforcement learning (RL), or imitation learning. 
The integration of semantic skills and the reasoning capabilities of LMs enables the zero-shot adaptation of EIF agents, without additional environment interactions or data collection efforts~\cite{llmplanner, progprompt, saycan}. In these approaches, an LM is prompted with the semantics of pretrained skills $L_{\pi} = \{l: \pi_l \in \Pi \}$ along with a specific instruction $i$ from the user. Subsequently, the LM produces a series of semantic skills. The execution of these skills facilitates the decomposition of the complex task associated with instruction $i$ into smaller, achievable skill-level subtasks.

\begin{figure*}[t]
    \centering
    \includegraphics[width=2.0\columnwidth]{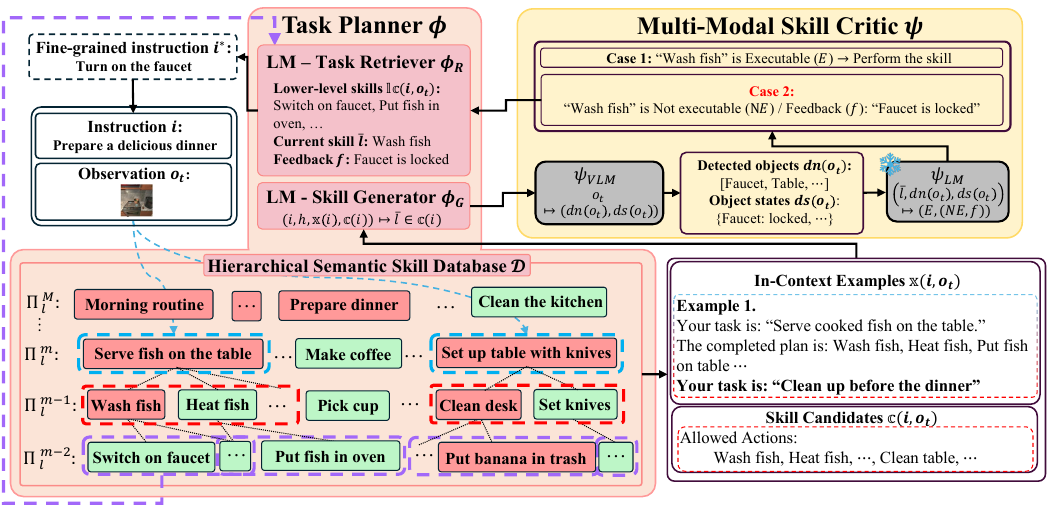}
    \caption{Overview of $\oursol$. Given a user instruction $i$ and detected object names from visual observations $o_t$, $\oursol$ selects $k$ entries (marked in blue dotted entries) from the hierarchical skill database $\mathcal{D}$. Within $\mathcal{D}$, skills are labeled based on their executability in the target domain: non-executable skills in red boxes and executable ones in green.
    From these selections, in-context examples $\mathbf{x}(i, o_t)$ and skill candidates $\mathbf{c}(i, o_t)$ are obtained from the constituent low-level skills (marked in red dotted entries in $\mathcal{D}$) of these entries. 
    The skill generator $\phi_G$ is then prompted with $\mathbf{x}(i, o_t)$, $\mathbf{c}(i, o_t)$, $i$, and skill execution history $h$ to generate the skill $\bar{l} \in \mathbf{c}(i, o_t)$.
    The LM-part $\psi_{\text{LM}}$ of the critic $\psi$ assesses the feasibility of the planned skill $\bar{l}$ based on the detected objects $dn(o_t)$ and their states $ds(o_t)$ as identified by the VLM-part $\psi_{\text{VLM}}$. 
    If $\bar{l}$ is deemed executable (Case 1), the agent proceeds to interact with the environment by performing the skill.
    Otherwise (Case 2), the task retriever $\phi_R$ generates a more fine-grained instruction $i^*$, derived from lower-level skills $\mathbf{lc}(i, o_t)$ (marked in purple dotted entries in $\mathcal{D}$) and feedback $f$ explaining the inexecutability. The $i^*$ is treated as a new instruction, prompting the framework to iterate the process.
    }
    \label{fig:fig2}
\end{figure*}

\subsection{Cross-Domain Skill Grounding}
While several works have explored the use of LMs as task planners, they primarily focus on single-domain agent deployment, wherein skills are both acquired and used within the same environment setting. 
In contrast, the cross-domain setting presents unique challenges for embodied task planning, as the feasibility of skill execution in the environment differs, necessitating the adaptation of pretrained skills to the specific domain contexts of the target environment. Consider the skill ``Make a cup of coffee'', as illustrated in Figure~\ref{fig:fig1}, learned in a high-tech kitchen environment equipped with voice-activated appliances, a digital inventory system, and automated utensil handlers. 
In this environment, the skill entails complex interactions with technology to efficiently prepare a cup of coffee. 
However, when a robot is deployed in a traditional kitchen devoid of such advanced features, it encounters a significantly different domain context. The skill ``Make a cup of coffee'', initially tailored for the environment with high-tech gadgets, now faces significant execution challenges due to the unavailability of similar technological supports.

To address this gap, we introduce \textbf{cross-domain skill grounding}, a process aimed at adapting pretrained skills $\pi_l \in \Pi$ for diverse domains. This process involves not only reinterpreting the original skill but also deconstructing it into a series of achievable lower-level subtasks, such as ``Boil water'', ``Add coffee to cup'', ``Pour water'', and ``Stir the drink''. 
Through cross-domain skill grounding, we ensure that embodied agents can perform a vast set of skills within domains that differ significantly from those in which the skills were initially learned. 

\section{Approaches}

We utilize the intrinsic semantic hierarchy of skills, a structure derived from their compositional nature; high-level semantic skills are established by chaining short-horizon low-level skills.
High-level skills, characterized by rich semantics,  closely align with the abstract nature of user instructions. This alignment in abstraction levels
facilitates the task planning process, providing the LM with skills that are directly beneficial for interpreting and accomplishing the instructions.
However, the execution of these long-horizon high-level skills is contingent upon the precise alignment of various environmental and domain-specific conditions. This  precludes their adaptability in cross-domain settings.
In contrast, low-level skills exhibit a robust capacity for execution that is not as tightly bound to specific domain contexts, offering greater adaptability across diverse domains. Nonetheless, the sheer volume of low-level skills and their limited semantic depths pose additional challenges for task planning. 

The core of $\oursol$ lies in the hierarchical task planning scheme via iterative skill grounding. 
%
The framework initiates task planning with high-level rich-semantic skills, and then progressively decomposes each of these planned high-level skills into a series of low-level skills executable within the domain of the target environment. 
This can be seen as finding an optimal intermediate semantic representation for skills pertinent to a given instruction, exploring a balance between their logical relevance and actual feasibility in the target environment. Such a balance ensures that the planned skills not only match the conceptual demands of the given instruction but are also effective within the limitations and resources of the target environment.

\subsection{Overall Framework}
As illustrated in Figure~\ref{fig:fig2}, $\oursol$ is structured with (a) a task planner and (b) a multi-modal skill critic, which collaboratively ground skills for user instructions in cross-domains. At each skill-step, the task planner produces an appropriate semantic skill by prompting an LM with information retrieved from a skill database. The skill critic then evaluates each skill from the skill planner by prompting a multi-modal LM with visual observations from the environment. When evaluated as unsuitable, the task planner decomposes the skill into smaller, more general subtasks using another retrieval from the database. Through this iterative process, the framework systematically assembles a sequence of semantic skills, executable within the domain context of the target environment.

\subsection{LM-based Task Planner}
\label{subsec:taskplanner}
For LM-based task planners, the development of effective prompts is essential to fully harness the reasoning capabilities of LMs. We consider two aspects: (i) selecting skill candidates that facilitate the skill-level task decomposition and (ii) incorporating in-context examples applied to similar instructions. 
To capture these aspects, we construct a hierarchical semantic skill database, leveraging the reasoning prowess of LMs. Integrated with a kNN retriever, this database not only assists in identifying suitable high-level skill candidates for task planning but also facilitating the retrieval of pertinent in-context examples that can further aid the planning process.
The task planner, enriched with this selection of skill candidates and in-context examples, ensures the application of suitable skill semantics in cross-domains.

\subsubsection*{Hierarchical semantic skill database.}
For the hierarchical semantic skill database $\mathcal{D}$, we use a structured approach to semantic skill training, similar to~\cite {boss}. 
The skill acquisition initiates with a set of short-horizon basic low-level skills $\Pi^1_l = \{\pi_l^{(1, 1)}, \cdots, \pi_l^{(1, N_1)}\}$ in the training environment, which can be acquired via RL or imitation learning. Under the LM guidance, the agent produces skill chains from $\Pi^1_l$, achieving high-level long-horizon skills $\Pi_l^2$. This bootstrapping process iteratively expands to form a comprehensive set of skill levels $\Pi_l = \{\Pi^1_l, \cdots, \Pi^M_l\}$.
This collection, encompassing skills from low-level to high-level, is used to establish the skill database $\mathcal{D}$:
\begin{equation}
    \label{eq: database}
    \begin{cases}
    \Pi^m_l = \{ \pi_l^{(m, 1)}, \cdots, \pi_l^{(m, N_m)}\} \\
    \mathcal{D} := \{e^{(m, n)}: 1 \leq m \leq M, 1 \leq n \leq N_m\} \\
    e^{(m, n)}:= (l^{(m, n)}, dn(e^{(m, n)}), p(e^{(m, n)})) \\
    l^{(m, n)}:= \text{Skill semantic of }\pi_l^{(m, n)}\\
    p(e^{(m, n)}) := (e^{(m-1, n_{j_1})}, \cdots).
    \end{cases}
\end{equation}
The entry $e^{(m,n)}$ of $\mathcal{D}$ includes 
the skill semantic $l^{(m,n)}$ of $\pi_l^{(m, n)}$,
detected object names $dn(e^{(m, n)})$ collected during training of $\pi_l^{(m, n)}$,
and a one-step low-level semantic skill planning $p(e^{(m, n)}) = (e^{(m-1, n_{j_1})}, \cdots)$, in that the skill $\pi_l^{(m, n)}$ is acquired by chaining of $(\pi_l^{{(m-1, n_{j_1})}}, \cdots )$.
To train the skill critic (Section~\ref{subsec: critic}), we also establish a dataset: 
\begin{equation}
    \label{eq: observation dataset}
    \mathcal{D}_O = \{o_t^{(TR)}, dn(o_t^{(TR)}), ds(o_t^{(TR)}): t\}
\end{equation}
of visual observations $o_t^{(TR)}$ in the training environment, detected object names $dn(o_t^{(TR)})$ and object physical states $ds(o_t^{(TR)})$ at each timestep $t$ during the establishment of $\Pi_l$. 
To obtain such object names (e.g., ``microwave'') and physical states (e.g., ``microwave: closed''), we use an open-vocabulary detector tailored for the training environment. 


%
%
\subsubsection*{Skill generator.}
For a given user instruction $i$ and target environment observation $o_t$, the LM-based skill generator $\phi_G$ integrates skill candidates $\mathbf{c}(i, o_t)$, in-context examples $\mathbf{x}(i, o_t)$, and the skill execution history $h$ to generate the skill semantic $\bar{l}$ which is most likely to be executed at the present:
\begin{equation}
    \label{eq: skill generator}
    \phi_G: (i, h, \mathbf{x}(i, o_t), \mathbf{c}(i, o_t)) \mapsto \bar{l} \in \mathbf{c}(i, o_t).
\end{equation}
To derive these in-context examples and skill candidates, we use a kNN retriever that selects entries $\mathbf{e}(i, o_t) = \{e^{(m_1, n_1)}, \cdots, e^{(m_k, n_k)}\}$ in  $\mathcal{D}$. 
The retrieval process calculates similarity by combining the instruction $i$ and the visual context of $o_t$ into a single query. It evaluates $\langle i, l^{(m, n)} \rangle + \langle dn(e^{(m, n)}), o_t \rangle$ to determine the $k$ entries with the highest similarity scores, where $\langle \cdot , \cdot \rangle$ denotes the cosine similarity. Here,
$l^{(m, n)}$ is the skill semantic and $dn(e^{(m, n)})$ is the detected object names associated with entry $e^{(m, n)}$.
%
%
These retrieved entries are then used to define one-step lower-level plans as in-context examples $\mathbf{x}(i, o_t)$ and one-step lower-level skill semantics $\mathbf{c}(i, o_t)$ as the planning candidates for the skill generator $\phi_G$:
\begin{equation}
    \label{eq: db exc}
    \begin{cases}
        \mathbf{e}(i, o_t) = \{(e^{(m_1, n_1)}, \cdots, e^{(m_k, n_k)})\} \\
        \mathbf{x}(i, o_t) = \{(l^{(m_1, n_1)}, p_l(e^{(m_1, n_1)}), \cdots \} \\
        \mathbf{c}(i, o_t) = \{l^{(m_1-1, n_{j_1})}, \cdots \}.
    \end{cases}
\end{equation}
Here, $p_l(e^{(m, n)})$ is the semantics in $p(e^{(m, n)})$.
%

\subsubsection*{Task retriever.}
If the planned skill semantic $\bar{l}$ from the skill generator $\phi_G$ is identified as inexecutable by the skill critic, the LM-based task retriever $\phi_R$ steps in to generate a detailed, fine-grained language instruction $i^*$. 
This new instruction $i^*$ is generated in response to the feedback $f$ regarding the rationale for this inexecutability and lower-level skill semantics $\textbf{lc}(i, o_t)$ that are constituent of the original skill semantic candidates $\mathbf{c}(i, o_t)$ in Eq~\eqref{eq: db exc}:
\begin{equation}
    \label{eq: task retriever}
    \begin{cases}
        \phi_R: (\bar{l}, f, \mathbf{lc}(i, o_t)) \mapsto i^* \\
        \mathbf{lc}(i, o_t) = \bigcup_{j=1}^k \{p_l(e'): e' \in p(e^{(m_j, n_j)}) \}.    
    \end{cases}
\end{equation}
This instruction $i^*$ guides the framework to perform  smaller, more manageable subtasks by triggering the reiteration of skill generation in Eq~\eqref{eq: skill generator}. 
%
\subsection{Multi-Modal Skill Critic}
\label{subsec: critic}
For observation $o_t$ and skill semantic $\bar{l}$ derived from the skill generator $\phi_G$ (in Eq~\eqref{eq: skill generator}), the skill critic $\psi$ assesses the execution feasibility of skill $\pi_{\bar{l}}$:
\begin{equation}
    \label{eq: skill critic}
    \psi: (o_t, \bar{l}) \mapsto c \in \{\text{E}, (\text{NE}, f)\}.
\end{equation}
Here, \text{E} specifies that the skill is executable, while \text{NE} denotes inexecutability with language feedback $f$. 
%
%
The critic $\psi$ includes the vision-language model (VLM)-part $\psi_{\text{VLM}}$ and LM-part $\psi_{\text{LM}}$. 
Here, $\psi_{\text{VLM}}$ identifies object names $dn(o_t)$ and states $ds(o_t)$ from $o_t$, and $\psi_{\text{LM}}$ utilizes this information to determine the executability of the skill $\pi_{\bar{l}}$:
\begin{equation}
    \label{eq: critic vlm lm}
    \begin{cases}
        \psi = \psi_{\text{LM}} \circ \psi_{\text{VLM}}: (o_t, \bar{l}) \mapsto \{\text{E}, (\text{NE}, f)\} \\
        \psi_{\text{VLM}}: o_t \mapsto (dn(o_t), ds(o_t)) \\
        \psi_{\text{LM}}: (\bar{l}, dn(o_t), ds(o_t)) \mapsto \{\text{E}, (\text{NE}, f)\}.
    \end{cases}
\end{equation}
We use the dataset $\mathcal{D}_O$ (in Eq~\eqref{eq: observation dataset}) to fine-tune the physically grounded VLM~\cite{pginstblip}. 
%
Our ablation study (Section~\ref{sec: ablation studies}) specifies the benefits of this LM's multi-modal extension with VLMs in cross-domains.  

\begin{algorithm}[t]
    \caption{Iterative Skill Grounding for EIF}
    \begin{algorithmic}[1]
    \STATE User instruction $i$, skill database $\mathcal{D}$
    \STATE Task planner $\phi = (\phi_G, \phi_R)$
    \STATE Skill critic $\psi = \psi_{\text{LM}} \circ \psi_{\text{VLM}}$
    \STATE Instruction stack $I \leftarrow [i]$
    \STATE Execution history  $H \leftarrow []$
    \STATE $t \leftarrow 0$, $o_t \leftarrow env\text{.reset()}$
    \REPEAT
        \STATE $i \leftarrow I\text{.pop()}, h \leftarrow H\text{.pop()}$
        \STATE $\mathbf{e}(i, o_t) \leftarrow \text{$k$ entries}$ via kNN retriever
        \STATE $\mathbf{x}(i, o_t) \leftarrow \text{in-context examples (Eq~\eqref{eq: db exc}}$)
        \STATE $\mathbf{c}(i, o_t) \leftarrow \text{skill candidate (Eq~\eqref{eq: db exc}}$)
        \STATE $\bar{l} \leftarrow \phi_{G}(i, h, \mathbf{x}(i, o_t), \mathbf{c}(i, o_t))$ (Eq~\eqref{eq: skill generator})
        \STATE $c \leftarrow \psi(o_t, \bar{l})$ (Eq~\eqref{eq: skill critic})
        \IF {$c = \text{E}$}
            \STATE $I \leftarrow I\text{.append}(i)$, $H \leftarrow H\text{.append}(\bar{l})$
            \STATE $t \leftarrow t + 1$
            \STATE $o_t \leftarrow env\text{.step}(\bar{l})$
        \ELSIF {$c=(\text{NE}, f)$ }
            \STATE $\mathbf{lc}(i, o_t) \leftarrow \text{lower skills (Eq~\eqref{eq: task retriever}}$)
            \STATE $i^* \leftarrow \phi_R(\bar{l}, f, \mathbf{lc}(i, o_t)) \text{(Eq~\eqref{eq: task retriever}}$)
            \STATE $I \leftarrow I\text{.append}(i)$
            \STATE $I \leftarrow I\text{.append}(i^*)$
        \ENDIF
    \UNTIL{$env\text{.done()}$}
    \end{algorithmic}
    \label{alg: skill grounding}
\end{algorithm}

The iterative skill grounding scheme of $\oursol$ is illustrated in Algorithm~\ref{alg: skill grounding}.

\section{Experiments}


\subsubsection*{Environment.}
We use VirtualHome, a realistic household activity simulator and benchmark.
%
For EIF agents, 
we define $3949$ low-level skills (i.e., $\Pi_1$ in Section~\ref{subsec:taskplanner}) based on the combination of behaviors (e.g., ``grasp'') and objects (e.g., ``apple''). 
%
%
\begin{table*}[t]
    \centering
    \begin{adjustbox}{width=1.0 \textwidth}
    \begin{tabular}{cc||ccc||ccc||ccc||ccc}
    \specialrule{.1em}{.05em}{.05em}
     \multirow{2}{*}{\textbf{Instruction}} & \multirow{2}{*}{\begin{tabular}{c} \textbf{Cross} \\ \textbf{Domain} \end{tabular}} & \multicolumn{3}{c||}{\textbf{LLM-Planner}} & \multicolumn{3}{c||}{\textbf{SayCan}} & \multicolumn{3}{c||}{\textbf{ProgPrompt}} & \multicolumn{3}{c}{\textbf{SemGro}}\\   
     &  & \textbf{SR} & \textbf{CGC} & \textbf{Plan} & \textbf{SR} & \textbf{CGC} & \textbf{Plan} & \textbf{SR} & \textbf{CGC} & \textbf{Plan} & \textbf{SR} & \textbf{CGC} & \textbf{Plan} \\ \specialrule{.1em}{.05em}{.05em}
    \multirow{3}{*}{\begin{tabular}{c} \textbf{Abstract} \\ \textbf{Noun} \end{tabular}} & \textbf{OL} & 20.51\% & 28.20\% & 43.59\% & 30.77\% & 34.19\% & 38.75\% & 17.95\% & 29.49\% & 34.90\% & 53.85\% & 67.31\% & 70.51\% \\
     & \textbf{PA} & 20.69\% & 35.63\% & 31.03\% & 17.24\% & 25.86\% & 27.24\% & 27.58\% & 37.93\% & 46.38\% & 52.17\% & 68.84\% & 71.74\% \\
     & \textbf{RS} & 16.22\% & 22.97\% & 29.73\% & 18.93\% & 32.48\% & 39.74\% & 13.51\% & 25.68\% & 29.28\% & 56.52\% & 73.91\% & 85.51\% \\ \specialrule{.1em}{.05em}{.05em}
     \multirow{3}{*}{\begin{tabular}{c} \textbf{Abstract} \\ \textbf{Verb} \end{tabular}} & \textbf{OL} & 28.57\% & 38.10\% & 45.24\% & 26.19\% & 40.48\% & 46.03\% & 28.57\% & 39.29\% & 39.29\% & 53.57\% & 67.86\% & 70.04\% \\
     & \textbf{PA} & 25.93\% & 44.44\% & 55.86\% & 19.23\% & 26.28\% & 33.07\% & 40.74\% & 54.32\% & 54.94\% & 61.54\% & 75.64\% & 80.45\% \\
     & \textbf{RS} & 30.95\% & 44.84\% & 39.81\% & 23.81\% & 37.70\% & 35.58\% & 35.71\% & 51.19\% & 40.61\% & 57.14\% & 71.43\% & 80.16\% \\ \specialrule{.1em}{.05em}{.05em}
     \multirow{3}{*}{\textbf{Structured}} & \textbf{OL} & 30.77\% & 35.90\% & 39.89\% & 38.89\% & 41.20\% & 46.30\% & 23.08\% & 30.77\% & 38.46\% & 51.28\% & 64.10\% & 69.23\% \\
     & \textbf{PA} & 25.93\% & 46.30\% & 48.89\% & 26.09\% & 39.13\% & 50.94\% & 15.38\% & 26.92\% & 44.87\% & 46.15\% & 60.26\% & 61.86\% \\
     & \textbf{RS} & 38.46\% & 42.31\% & 49.72\% & 25.64\% & 46.58\% & 45.44\% & 21.05\% & 30.26\% & 38.16\% & 43.59\% & 61.11\% & 70.51\% \\ \specialrule{.1em}{.05em}{.05em}
     \multirow{3}{*}{\begin{tabular}{c} \textbf{Long} \\ \textbf{Horizon} \end{tabular}} & \textbf{OL} & 25.81\% & 40.32\% & 48.92\% & 41.94\% & 46.24\% & 46.59\% & 35.48\% & 43.55\% & 53.58\% & 54.84\% & 64.52\% & 68.81\% \\
     & \textbf{PA} & 29.63\% & 58.64\% & 55.80\% & 40.74\% & 70.37\% & 58.09\% & 44.44\% & 64.81\% & 57.22\% & 62.96\% & 74.07\% & 61.11\% \\
     & \textbf{RS} & 25.93\% & 39.51\% & 38.27\% & 35.48\% & 46.77\% & 41.22\% & 30.95\% & 48.81\% & 38.22\% & 51.61\% & 60.21\% & 68.64\% \\ 
    \specialrule{.1em}{.05em}{.05em}
    \end{tabular}
    \end{adjustbox}
    \caption{Cross-domain performance in VirtualHome. The success rate (\textbf{SR}), completed goal conditions (\textbf{CGC}), and planning accuracy (\textbf{Plan}) are measured with 3 different seeds for each cross-domain scenario.}
    \label{table:main results}
\end{table*}
For cross-domain evaluations, we use 15 distinct environment settings, categorized by one of domain shift types: 
\textbf{OL} involves changes in the placements and locations of objects; 
\textbf{PA} presents changes in the physical attributes of objects; 
\textbf{RS} is characterized by different room structures and visual attributes.
These cross-domain contexts necessitate the refinement of certain pretrained high-level skills into low-level skills for successful execution. Consider a high-level skill ``cook salmon'' intended to fulfill the user instruction ``prepare delicious food for me''. If this skill was trained in a setup where salmon is always prepared on a cutting board, it might not work in \textbf{OL} cross-domain scenarios where salmon is instead found inside a refrigerator. In this case, it becomes necessary to decompose the skill into a sequence of lower-level executable skills such as ``go to the refrigerator'', ``grasp the refrigerator handle'', among others, ensuring the skill be adapted to the new domain context.


%

\subsubsection*{Instructions.}
Given 5 evaluation tasks, we use 4 different instruction types per task: \textbf{Abstract Nouns} and \textbf{Abstract Verbs} focus on instruction queries involving abstract nouns and verbs, respectively; \textbf{Structured} language queries include precise verbs; \textbf{Long-Horizon} queries involve multiple reasoning steps. These instruction types are used to test various linguistic aspects of task interpretation and execution, similar to~\cite{saycan}. 
By combining 15 environment settings, 5 tasks, and 4 types of instructions,  we establish a comprehensive set of 300 distinct unseen cross-domain scenarios. 

\subsubsection*{Evaluation metrics.}
We use several evaluation metrics, consistent with previous works~\cite{progprompt, llmplanner}. \textbf{SR} measures the percentage of successfully completed tasks, where a task is considered to be completed if its all goal conditions are completed; \textbf{CGC} measures the percentage of completed goal conditions; \textbf{Plan} measures the percentage of the planned skill sequence that continuously aligns with the ground-truth sequence from the beginning.

\subsubsection*{Baselines.}
We compare our $\oursol$ with three baselines, each leveraging an LM as a task planner but with distinct approaches to filter executable skill candidates, including
%
\textbf{SayCan}~\cite{saycan} which employs an offline RL dataset collected from the training environment to learn an affordance function; 
%
\textbf{LLM-Planner}~\cite{llmplanner} which leverages templatized semantic skills; 
%
\textbf{ProgPrompt}~\cite{progprompt} which uses human-crafted programmatic assertion syntax to verify the pre-conditions of skill execution and respond to the failed cases with recovery.

\subsection{SemGro Implementation}
For the kNN retriever, we configure $k$ to be 10 and utilize the sentence-transformer~\cite{sentence_transformer} to compute the similarity between the instruction and skill semantics in the database. 
The task planner $\phi$ and the LM-part $\psi_{\text{LM}}$ of the skill critic are both implemented using GPT-3.5 API with the temperature hyperparameter set to $0$. For fair comparisons, the same GPT-3.5 configuration is used across all baseline task planners. More details, including the baseline implementation, are in the Appendix.

\subsection{Performance}
\subsubsection*{Overall performance.}

Table~\ref{table:main results} compares the cross-domain performance, achieved by our $\oursol$ and the baselines (LLM-Planner, SayCan, ProgPrompt). 
$\oursol$ outperforms the baselines in all cross-domain scenarios, achieving average performance gains of 
$25.02\%$, $26.83\%$, and $27.65\%$ over the most competitive baseline SayCan in \textbf{SR} and \textbf{CGC}, and LLM-Planner in \textbf{Plan}, respectively. 
The exceptional performance of $\oursol$ highlights its superiority in hierarchical task planning via iterative skill grounding across unseen domains.  


Interestingly, for long-horizon instruction scenarios, there are several instances where \textbf{SR} and \textbf{CGC} metrics exceed \textbf{Plan}. In the 
\textbf{PA} cross-domain context, $\oursol$ achieves \textbf{Plan} of 61.11\%, while \textbf{SR} and \textbf{CGC} reach 62.96\% and 74.07\%, respectively. 
This phenomenon, reminiscent of solving math problems in an NLP context~\cite{step_by_step, fu_math, mathprompter}, suggests that tasks requiring multiple reasoning steps can be accomplished through diverse skill sequences.

\subsubsection*{Executable skill identification.}
We investigate the efficacy of $\oursol$ in generating executable skills. To this end, we introduce the \textbf{Exec} metric, which quantifies the percentage of the planned semantic skills that are executable in a given domain, regardless of their impact on task completion.
In Table~\ref{table: skill identifiability}, $\oursol$  surpasses the most competitive baseline ProgPrompt by $17.99\%$ in \textbf{Exec} at average. This specifies our framework's proficiency in discerning the varying conditions of cross-domain environments for skill grounding. This ability is attributed to the skill critic which uses the domain-agnostic representation capabilities of the adopted VLM.
\begin{table}[h]
    \centering
    \resizebox{\columnwidth}{!}
    {
        \begin{tabular}{c||c|c|c|c}
    	\specialrule{.1em}{.05em}{.05em}
        \multirow{2}{*}{\begin{tabular}{c} \textbf{Cross} \\ \textbf{Domain} \end{tabular}} 
        & \multirow{2}{*}{\textbf{LLM-Planner}}
        & \multirow{2}{*}{\textbf{SayCan}} 
        & \multirow{2}{*}{\textbf{ProgPrompt}} 
        & \multirow{2}{*}{\textbf{SemGro}} \\
        & & & & \\
        \specialrule{.1em}{.05em}{.05em}
        \textbf{OL} & 40.28\% & 60.22\% & 68.29\% & 85.05\% \\ 
        \textbf{PA} & 46.46\% & 68.18\% & 72.55\% & 97.40\% \\ 
        \textbf{RS} & 43.48\% & 69.41\% & 67.74\% & 79.12\% \\ 
        \specialrule{.1em}{.05em}{.05em}
        \end{tabular}
    }
    \caption{Executable skill identification}
    \label{table: skill identifiability}
\end{table}
%
%

\subsubsection*{Iterations for cross-domain skill grounding.}
%
We investigate the number of iterations used for skill grounding in the target domain across varying degrees of domain shift. 
To quantify the degree of shift, we calculate the number of high-level skills that must be decomposed into different series of low-level skills for execution in the target domain, compared to the training environment.
The degree of shift is categorized into 4 levels. `None' indicates no domain shift, while `Small,' `Medium,' and `Large' denote increasing degrees of shift. 
The detailed criterion for quantifying domain shift degrees can be found in Appendix B.
To accurately measure the iterations induced by domain shift, distinctions are made between skills rendered inexecutable due to the immediate observational state (Obs. \& Dom. in the row name) and those affected by domain disparities (Dom.).
Table~\ref{table: iteration frequency} reveals a positive correlation between the domain shift degrees and the number of iterations. This specifies the benefits of $\oursol$ in handling cross-domain settings with varying degrees of domain shift, progressively searching the appropriate abstraction level of skills.


\begin{table}[h]
    \centering
    \resizebox{\columnwidth}{!}
    {
        \begin{tabular}{c||c|c|c|c}
    	\specialrule{.1em}{.05em}{.05em}
        & \textbf{None} & \textbf{Small} & \textbf{Medium} & \textbf{Large} \\
        \specialrule{.1em}{.05em}{.05em}
        \textbf{Obs. \& Dom.} & 3.73 & 5.53 & 5.80 & 6.53 \\
        \textbf{Dom.} & 0.87 & 1.64 & 1.96 & 2.64 \\

        \specialrule{.1em}{.05em}{.05em}
        \end{tabular}
    }
    \caption{The number of iterations w.r.t. domain shift}
    \label{table: iteration frequency}
\end{table}

\subsection{Ablation Studies}
\label{sec: ablation studies}

\subsubsection*{Skill hierarchy for planning.}
To verify the effectiveness of the iterative skill grounding scheme, we conduct task planning with the fixed abstraction level of skills, bypassing the executability check of the skill critic.
In Table~\ref{table: lowest highest}, SG-L and SG-H employ the lowest-level skills $\Pi_l^1$ and highest-level skills $\Pi_l^M$ for task planning, respectively, while SG-M employs middle-tier skills $\{\Pi_l^m: 1<m<M\}$.
%
As shown, SG-H exhibits the most accurate task planning (high \textbf{Plan}), while its planned skills are rarely executable (low \textbf{Exec}) due to domain shift. 
Conversely, SG-L, equipped with potentially executable skills, shows unreliable task planning due to the abstraction level mismatch with the instruction.
Although SG-M employs skills with moderate semantic content and executability by adhering to a fixed abstraction level, it falls short in performance. 
$\oursol$ adaptively identifies skill semantics at the intermediate level, through  iterative downward decomposition within the semantic hierarchy of skills.
This strategy not only facilitates task planning but also achieves robust executions in new domains.
%


\begin{table}[h]
\centering
\resizebox{\columnwidth}{!}{
\begin{tabular}{cc||c|c|c|c}
\specialrule{.1em}{.05em}{.05em}
\multicolumn{2}{c||}{\multirow{2}{*}{\begin{tabular}{c}\textbf{Cross}\\ \textbf{Domain}\end{tabular}}} & \multirow{2}{*}{\textbf{SG-L}} & \multirow{2}{*}{\textbf{SG-M}} & \multirow{2}{*}{\textbf{SG-H}} & \multirow{2}{*}{\textbf{SemGro}} \\
& & & & & \\
\specialrule{.1em}{.05em}{.05em}
\multirow{2}{*}{\textbf{OL}} & \multirow{2}{*}{\begin{tabular}{c}\textbf{Plan}\\ \textbf{Exec}\end{tabular}} & 46.30\% & 65.93\% & 89.26\% & 82.96\% \\
& & 93.05\% & 75.64\% & 56.15\% & 92.27\% \\
\hline
\multirow{2}{*}{\textbf{PA}} & \multirow{2}{*}{\begin{tabular}{c}\textbf{Plan}\\ \textbf{Exec}\end{tabular}} & 48.45\% & 69.78\% & 80.56\% & 78.44\% \\
& & 91.80\% & 78.76\% & 57.32\% & 91.02\% \\
\hline
\multirow{2}{*}{\textbf{RS}} & \multirow{2}{*}{\begin{tabular}{c}\textbf{Plan}\\ \textbf{Exec}\end{tabular}} & 35.56\% & 48.15\% & 87.04\% & 85.19\% \\
& & 89.72\% & 73.93\% & 44.92\% & 87.03\% \\
\specialrule{.1em}{.05em}{.05em}
\end{tabular}
}
\caption{Skill hierarchy for planning}
\label{table: lowest highest}
\end{table}


\subsubsection*{In-context example retrieval.}
In Table~\ref{table: in-context}, we conduct an ablation study on the kNN retriever 
by varying the number $k$ of retrieved entries in Eq~\eqref{eq: db exc}. Furthermore, we test a random-selection retriever that selects 10 examples at random (Random in the column name) as well as another variant that calculates the similarity based solely on instructions without observation information ($\mathbf{k=10}$ $\setminus O$ in the column name). 
For the random-selection retriever, skill candidates are determined using the kNN retriever to precisely assess the impact of in-context examples.
As observed, selecting an appropriate number of examples ($k=10$) leads to performance gain in \textbf{Plan} by 30.89$\%$ and $7.87\%$ compared to the random retriever and $\mathbf{k=10}$ $\setminus O$, respectively. 
Furthermore, we observe an increase in planning performance when $k$ ranges from $5$ to $10$, but beyond this range, larger values do not yield further enhancements. 
This implies that including skill candidates unrelated to the task may deteriorate the task planning.
\begin{table}[h]
    \centering
    \resizebox{\columnwidth}{!}
    {
        \begin{tabular}{c||c|c|c|c|c}
        \specialrule{.1em}{.05em}{.05em}
        \multirow{2}{*}{\begin{tabular}{c} \textbf{Cross} \\ \textbf{Domain} \end{tabular}} 
        & \multirow{2}{*}{\begin{tabular}{c} \textbf{Random} \\ ($\mathbf{k=10}$) \end{tabular}} 
        & \multirow{2}{*}{$\mathbf{k = 5}$} 
        & \multirow{2}{*}{$\mathbf{k = 10}$}
        & \multirow{2}{*}{$\mathbf{k = 15}$}
        & \multirow{2}{*}{$\mathbf{k=10} \setminus O $}\\
        & & & & & \\
        \hline
        \textbf{OL} & 29.63\% & 48.51\% & 69.63\% & 47.04\%  & 59.25\% \\ 
        \textbf{PA} & 65.56\% & 61.25\% & 81.94\% & 79.86\%  & 80.56\% \\ 
        \textbf{RS} & 34.07\% & 58.15\% & 70.37\% & 64.81\%  & 58.52\% \\ 
        \specialrule{.1em}{.05em}{.05em}
    	\end{tabular}
    }
    \caption{In-context example retrieval}
    \label{table: in-context}
\end{table}

\subsubsection*{The dual-model structure for skill critic.} 
As explained in Section~\ref{subsec: critic}, we structure the skill critic $\psi$ with two models $\psi_{\text{VLM}}$ and $\psi_{\text{LM}}$. To validate this dual-model structure, we conduct an ablation study using a single VLM for the skill critic. Specifically, we use InstructBLIP~\cite{instblip}, a state-of-the-art general-purpose VLM, alongside its fine-tuned counterpart PG-InstructBLIP~\cite{pginstblip}, which is optimized for object-centric physical understanding.
We fine-tune these VLMs on the VirtualHome dataset. We provide the dataset collection process in the Appendix.
%
%
In Table~\ref{table: unified vlm}, the dual-model structure for $\psi$ improves the performance in \textbf{Exec} by $9.62\%$ over the singular VLM. This specifies that while VLMs are adept at assessing physical attributes, their capacity to discern the executability of semantic skills in diverse physical environments can be more effectively augmented by the inclusion of LMs.
\begin{table}[h]
    \centering
    \resizebox{\columnwidth}{!}
    {
        \begin{tabular}{c||c|c|c}
        \specialrule{.1em}{.05em}{.05em}
        \multirow{2}{*}{\begin{tabular}{c} \textbf{Cross} \\ \textbf{Domain} \end{tabular}}  
        & \multirow{2}{*}{\textbf{$\psi_{\text{VLM}}$-IB}} 
        & \multirow{2}{*}{\textbf{$\psi_{\text{VLM}}$-PGIB}}
        & \multirow{2}{*}{\textbf{$\psi_{\text{LM}} \circ \psi_{\text{VLM}}$}} \\
        & & & \\
        \hline
        \textbf{OL} & 79.67\% & 77.36\% & 89.06\%  \\ 
        \textbf{PA} & 87.32\% & 91.53\% & 98.46\%  \\ 
        \textbf{RS} & 70.59\% & 71.15\% & 81.39\%  \\ 
        \specialrule{.1em}{.05em}{.05em}
    	\end{tabular}
    }
    \caption{The dual-model structure for skill critic}
    \label{table: unified vlm}
\end{table}

\subsubsection*{LMs for task planner and skill critic.}
To implement the task planner $\phi = (\phi_G, \phi_R)$ and the LM-part $\psi_{\text{LM}}$ of the skill critic, we test a range of LMs from open-source LLaMA-2-70B to proprietary capable LMs such as PaLM, GPT-3.5, and GPT-4. 
In Table~\ref{table: llm for task planner}, while LLaMa-2-70B, with significantly fewer parameters, does not achieve comparable performance, $\oursol$ demonstrates robust performance consistently with capable LM backbones.

\begin{table}[h]
    \centering
    \resizebox{\columnwidth}{!}
    {
        \begin{tabular}{c||c|c|c|c}
    	\hline
    	\textbf{Metric} & \textbf{LLaMa-2-70B} & \textbf{PaLM} & \textbf{GPT-3.5} & \textbf{GPT-4} \\
        \hline
        \textbf{SR} & 25.93\% & 46.15\% & 46.43\%  & 48.15\% \\ 
        \textbf{CGC} & 43.20\% & 64.10\% & 62.25\%  & 60.49\% \\
        \textbf{Plan} & 50.43\% & 67.73\% & 66.09\%  & 67.90\% \\ 
        \textbf{Exec} & 71.21\% & 91.81\% & 89.76\%  & 85.63\% \\

        \hline
    	\end{tabular}
    }
    \caption{LMs for task planner and skill critic}
    \label{table: llm for task planner}
\end{table}

\section{Related Works}

\subsubsection*{Language models for embodied control.}
The use of pretrained LMs for embodied control tasks has been increasingly prevalent. A significant focus of this research is on bridging the gap between LMs' extensive knowledge and their lack of inherent understanding of embodied experiences. 
Several works tackle the challenge by directly incorporating environmental contexts to LMs through textual prompts~\cite{e2wm, gplanet, swiftsage, reflexion, glam}, based on the premise that observations can be described in text form.
For example,~\cite{gplanet} utilizes a text-based table to display objects' location and their relation. 
%
Expanding beyond the mere textual environment, other works have pivoted towards visual tasks and EIF~\cite{describe, llmplanner, palm-e, tapa, progprompt}.
Particularly,~\cite{saycan} introduces a method to learn skill affordance from visual observations using an offline RL dataset collected from the specific environment. This  approach integrates skill logits from an LM with identified skill affordance to select the most valuable skill for execution. 
Meanwhile,~\cite{tapa, llmplanner} use detected object classes in the observation to refine the selection of actionable skills.

Our $\oursol$ distinguishes itself in EIF by being the first to harness the capabilities of LMs for grounding pretrained skills in cross-domains. 
%

%


\subsubsection*{Language conditioned policy.}
A number of works have concentrated on the generalization of language-conditioned policies or skills for unseen language instructions~\cite{hiveformer, lisa}, visual attributes~\cite{rt1, rt2, owlvit, perceiver-actor, cliport}, or environment dynamics~\cite{onis}, with some extending to multi-modal instructions~\cite{vima2, mutex, semtra, vima, bcz}.
For instance,~\cite{rt2} adapts a pretrained foundation model to robot trajectory datasets in an end-to-end manner, by representing robot actions as text tokens. 
\cite{vima2, vima} offer a unified framework capable of achieving interleaved vision-language instructions, employing a set of pretrained models to encode them. 

While prior works emphasize the robust policy structure, $\oursol$ leverages an LM-based iterative reasoning mechanism, aiming for balanced skill grounding across diverse domain shifts. Moreover, $\oursol$ addresses realistic EIF tasks with the VirtualHome benchmark, unlike the prior works that often focus on straightforward manipulation tasks. 

\section{Conclusion}
In this paper, we presented a semantic skill grounding framework $\oursol$, which supports the adaptation of a broad range of skills in cross-domain environments. 
$\oursol$ harnesses the reasoning capabilities of LMs not only to compose semantic skills for establishing a hierarchical skill database but also to decompose semantic skills into executable low-level skills. 
This novel approach uniquely explores the trade-off between semantic richness and domain-agnostic executability of semantic skills. We hope our work provides valuable insights into the development and application of real-world EIF agents.

\section{Limitations}
Despite the strong performance of $\oursol$, we identify several limitations for improvement.
\textbf{(i) Single-step visual observation for skill feasibility:} 
Currently, the skill feasibility is assessed based on a single observation. This approach can be improved by integrating a VLM into our framework that is capable of generating imaginary observations, at the expense of increased resource usage by the embodied agent. 
By employing Monte Carlo Tree Search, the feasibility of skills could then be computed across multiple planning paths. Addressing this enhancement remains for our future work.

\textbf{(ii) Unidirectional decomposition strategy:} 
The strategy for semantic skill decomposition is unidirectional, proceeding only in a downward direction.
This may often restrict the framework as it cannot revert to initiating planning from a high-level skill at an appropriate juncture. 
Introducing bidirectionality in task planning could improve the versatility of $\oursol$. In scenarios with substantial domain shifts, such bidirectionality would allow for completely novel planning that has not been trained to achieve complex, unseen tasks. Enhancing our framework to incorporate this bidirectional planning presents a challenge but crucial direction for future development.

\section{Ethics Statement}
This work introduces a technology that can impact the application of robotics in household tasks. 
This technology signals a major shift in the human experience, as household tasks increasingly become automated by robotic agents. The deployment of robotic agents in household tasks presents both significant benefits and challenges. 
On the positive side, this technology can save time and effort for users, leading to increased convenience and efficiency in managing household chores.
However, it also raises important ethical considerations related to labor displacement. As robotic agents take over tasks traditionally performed by humans, there is a potential impact on the livelihoods of current service providers who may need to seek alternative sources of income. 
This transition necessitates careful consideration and planning to mitigate adverse effects on workers and to ensure that the benefits of technological advancement are equitably distributed. 
By acknowledging and addressing these societal impacts, our research aims to contribute positively to the field of robotics while fostering an inclusive approach that considers the broader implications for society.

\section{Acknowledgements}
We would like to thank anonymous reviewers for their valuable comments and suggestions.
This work was supported by Institute of Information \& communications Technology Planning \& Evaluation (IITP) grant funded by the Korea government (MSIT)
(No.
2022-0-01045, 
2022-0-00043, 
2019-0-00421, 
2020-0-01821),    
and by the National Research Foundation of Korea (NRF) grant funded by MSIT (RS-2023-00213118). 
\bibliography{custom}

\clearpage
\newpage
\appendix
\title{Appendix}
\section{Implementation Details}
In this section, we present the implementation details of each baseline method and our framework SemGro. 

\subsection{LLM-Planner}
The LLM-Planner~\cite{llmplanner} employs templatized semantic skills, k-nearest neighbors (kNN) retrievers, and a Language Model(LM) planner to generate plans for embodied tasks. The semantic skills are defined through templates which then are matched with the objects visible in the environment; this allows for instant skill construction options based on the specific situation observed. The LLM-Planner initially retrieves in-context examples from the training data, utilizing k-nearest neighbors (kNN) retrievers, which are then prompted in the LM planner. Subsequently, the planner merges these skill templates with currently visible objects to determine the  skill that is both achievable and capable of completing the task.

We use the open-source code from the llm-planner official github. Due to the lack of existing training datasets in VirtualHome, we exploit data from the primitive level of our skill database (in SemGro) as the training data for the retriever to generate in-context examples. We use GPT-3.5 turbo for the planner and a BERT-base-uncased model from the Huggingface Transformers library for the kNN retriever.

\subsection{SayCan}
SayCan~\cite{saycan} employs an LM planner, pre-defined semantic skills, and an affordance value function to generate a feasible skill plan for user instructions. The LM planner is responsible for finding the suitable skill from the skill set to achieve the given task using the LM log probability between the human instruction and the pre-defined set of skills. Then, the affordance score of each skill in the skill set is calculated using the pre-trained affordance value function. Saycan combines two scores and uses them to choose the appropriate feasible skill to achieve the task.  

We implement SayCan in a similar way as LLM-Planner. First, in-context examples are generated using kNN retrieval from our framework. The retrieved in-context examples are then prompted to the LM planner. We found that the skill generation is dependent on the in-context prompts. To facilitate the LM Planner in producing skills that are achievable in the current observations, we provide additional task examples that are attainable within the present context. Then, the planner chooses a suitable skill from the skill candidates to accomplish the task. We use GPT-3.5 turbo for the planner.

We use the formulation of skill and value functions similar to LLM-Planner. In SayCan, each skill is formulated as actions that can perform a short-horizon task, such as picking up a particular object. Similarly, we use the atomic actions provided in VirtualHome, such as grab, walk, switchon, and also combine them with individual objects to formulate the set of low-level skills. Subsequently, the value function emerges as another critical element in SayCan, evaluating the feasibility of each skill in light of the present observation. Due to the challenges associated with training low-level policies in VirtualHome, we adopt the approach used by LLM-Planner, providing SayCan with Oracle object data to establish the Oracle value function. This provides SayCan with preemptive access to comprehensive knowledge about all objects and their affordance in the current observation; thus, this shrinks the list of potential skills for the LLM to consider. Consequently, this simplifies the decision-making process for the planner, facilitating the selection of a suitable skill to complete the task effectively.

\subsection{ProgPrompt}
ProgPrompt~\cite{progprompt} employs a human-designed programming assertion syntax to check the pre-conditions for executing skills and addresses failures by initiating predefined recovery skills. In detail, ProgPrompt uses a plan structure consisting of comments, actions, and assertions in the style of programming language. Comments are used to cluster high-level skills together and actions are expressed as imported function calls. Assertions are used to check conditions right before actions start and execute the recovery skills if needed.

We use the code from the ProgPrompt official github. We use the same plan templates as employed by ProgPrompt, featuring a Pythonic style where the task name is designated as a function, available actions are included through headers, accessible objects are specified in variables, and each action is delineated as a line of code within the function. Similar to SayCan implementation, we also use knn retrieval to generate in-context examples for the LM Planner. We use GPT-3.5 turbo for the planner. 

Utilizing assertions to verify the pre-conditions of each action is a pivotal concept in ProgPrompt, enabling the correction of incorrect behaviors to accomplish tasks. However, generating pre-conditions for every action involves substantial effort and resources; furthermore, given the assumption that an innumerable set of skills could be integrated, manually annotating pre-conditions for each skill becomes practically infeasible. Similar to SayCan implementation, we endow ProgPrompt with Oracle pre-conditions for each action. By implementing Oracle assertions before each skill, the system automatically corrects behaviors, thus allowing the agent to perform only feasible skills to complete the task efficiently.

\subsection{SemGro}
Our SemGro framework comprises two primary components: the task planner $\phi$, which is composed of the skill generator $\phi_G$ and task retriever $\phi_R$, and the multi-modal skill critic $\psi$. We employ the kNN retriever to extract $k$ in-context examples from the hierarchical skill database using the instruction $i$ and current visual observations $o_t$. These $k$ in-context examples' low-level skill plans serve as candidates for the skill generator $\phi_G$. 

The execution feasibility of the planned skill is then assessed by the multi-modal skill critic $\psi$ based on the current observation. Within the critic, the Vision-Language Model (VLM)-part utilizes visual observations from the environment to infer detected objects and their states. Leveraging this inferred information, the LM-part of the critic assesses the execution feasibility of the planned skill. If the skill is assessed as executable, the skill is performed as it is; otherwise, we prompt the feedback from the critic along with the one-step further lower-level skill candidates to generate alternative, more fine-grained instructions. Utilizing the revised instructions, the database is queried for skills at a lower abstraction level to reinitiate the skill grounding process. An example of detailed process of our framework is in Table 10.

This iterative feedback mechanism ensures that the agent can adaptively refine its actions to better align with the dynamic conditions of its target physical environment, thereby enhancing the overall effectiveness and precision of skill execution.
In the development of our skill generator, we utilize GPT-3.5 Turbo. For the multi-modal critic, we employ InstructBLIP as the VLM-part, and GPT-3.5 as the LM-part. For the k-Nearest Neighbors (kNN) retriever, we integrate the BERT-base-uncased model from the Huggingface transformers library.  

\section{Experiment Details}
In this section, we describe the experiment settings in detail. We also specify the cross-domain scenarios and environment details used in the experiments, as well as the data collection process for our hierarchical skill database and the training procedure of our multi-modal critic.

\noindent\textbf{Cross-domain scenarios.}
We used 5 EIF tasks, 4 types of language instructions, and 15 environmental scenes to test across 300 cross-domain scenarios. The variation in instructions, akin to the approach used in SayCan, is crafted to assess the extent to which our framework can deconstruct abstract instructions and ground skills in the target domain. We utilize 4 instruction types such as Abstract Noun, Abstract Verb, Structured, and Long Horizon, comprising a total of 137 instructions.

The \textbf{Abstract Noun} and \textbf{Abstract Verb} categories are created to examine whether SemGro can execute the appropriate sequences when given abstractions are in nouns and verbs. \textbf{Structured Language}, paralleling Abstract Verbs, aims to evaluate SemGro's performance with well-structured languages. \textbf{Long Horizon}, necessitating multi-step reasoning, is designed to evaluate SemGro's capability to process and reason through temporally extended instructions.

\noindent\textbf{VirtualHome simulation.}
We use VirtualHome~\cite{virtualhome}, an interactive simulation platform for complex household tasks. The skills provided in the environment include manipulation actions such as picking up objects and toggling appliances on or off. We impose restrictions on these actions within the environment to align with our settings; the skills performed are implemented to be feasible only with the objects currently visible. For example, ``grab apple'' is executable when the apple is closed enough and is visible to the agent. The action space of the VirtualHome experiments is provided in Table 8.

\begin{table}[h]
\caption{VirtualHome Actions}
\begin{center}
\begin{small}
\begin{tabular}{ll}

\toprule
\multicolumn{2}{c}{\textbf{Actions}}\\
\midrule
 & find <obj>, grab <obj>, walk <obj>, sit <obj>  \\ 
 & put <obj> <obj>, open <obj>, close <obj> \\
 & switchon <obj>, switchoff <obj> \\
\bottomrule

\end{tabular}
\end{small}
\end{center}
\end{table}

We use a single training environment and 15 distinct environment settings for cross-domain evaluation, with the training environment being environment ID 0 in VirtualHome. We use three different domain-shift types, OL, PA, and RS as explained in the main manuscript. For each type, we use 5 distinct environment settings.
The OL scenario involves alterations in the locations of objects and their relationships compared to the training environment. 
The PA scenario reflects changes in the physical attributes of objects from the training environment. 
RS scenario encompasses shifts in the visual context, object locations, and their physical attributes.
Figure 3 presents visual examples of tasks, and Table~\ref{table: VH scenarios} details the cross-domain scenario examples employed in our experiments.



\begin{figure}
    \centering
    \includegraphics[width=1.0\columnwidth]{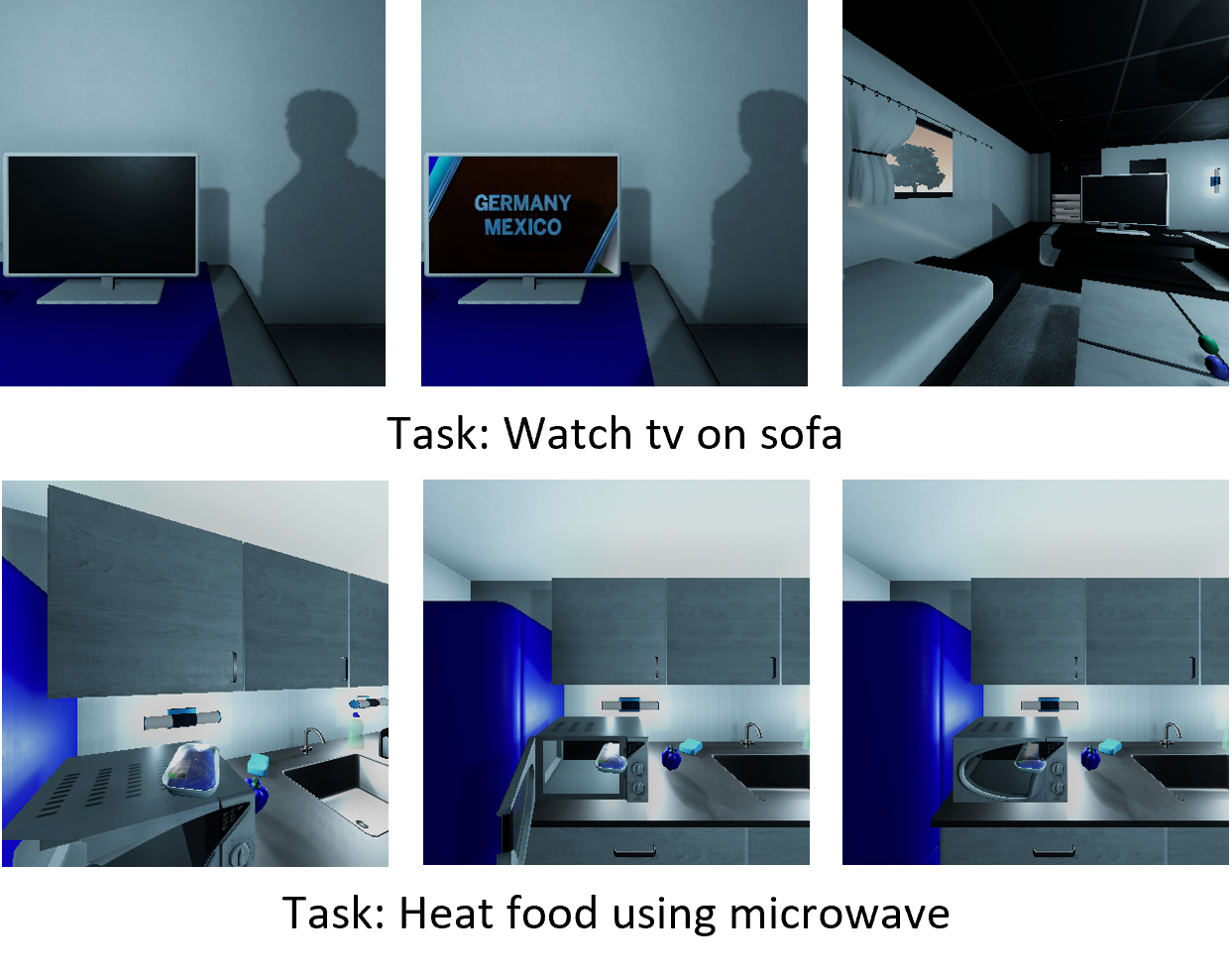}
    \caption{Caption}
    \label{Example tasks in VirtualHome}
\end{figure}

\noindent\textbf{Dataset collection process.}
The skill database $\mathcal{D}$ comprises 4 distinct abstraction levels, each containing 3949, 183, 142, and 90 skills, from the first to the fourth level, respectively.
We adopt a bottom-up approach to construct the skill database by progressively chaining low-level skills to form complex high-level skills. Specifically, we employ the skill bootstrapping method in~\cite{boss}, which leverages an LM to guide the development of high-level skills from pretrained low-level skills.

Upon initiation in the environment, the agent selects a skill from the available low-level skills. The LM is then queried to identify the most suitable next skill to execute, based on the provided prompt. Following this, the agent performs the selected skill. This cycle repeats until reaching a predetermined maximum number of steps or encountering a skill execution failure. 
Successful skill executions are summarized by the LM to derive a high-level abstract skill name, enabling the accumulation of a comprehensive skill set across different abstraction levels. For both skill selection and summarization, the text-bison-001 model from PaLM~\cite{palm} is utilized.

Furthermore, we use an open-vocabulary detection model owl-ViT~\cite{owlvit} to obtain the object names $dn(o^{(TR)})$ (e.g., ``microwave'') and InstructBLIP~\cite{instblip} to obtain the physical state of each object $ds(o^{(TR)})$ (e.g., ``microwave: closed'').

\noindent\textbf{VLM fine-tuning.}
We fine-tune the PG-InstructBLIP to implement the VLM-part $\psi_{\text{VLM}}$ of the skill critic, using the  dataset $D_O$ in Eq~\eqref{eq: observation dataset}. This process enhances the VLM's understanding of physical objects within the VirtualHome environment by adopting a Visual Question Answering (VQA) training approach. An object detector is employed to isolate the portions of images that feature objects, and these cropped images are then used to train the VLM in recognizing the objects' states. Specifically, we formulate questions about the objects' conditions (e.g., ``Is this object open or closed? Answer 'unknown' if uncertain. Short answer:'') and train the VLM to accurately answer these queries.

We use the FlanT5-XL~\cite{flant5} variant of InstructBLIP~\cite{instblip}, implemented in vision-language multi-modal open-source project LAVIS~\cite{lavis}. The model is trained for a total of 12 GPU hours on a system equipped with an Intel(R) Core(TM) i9-10980XE processor and an NVIDIA RTX A6000 GPU.

\noindent\textbf{Quantification of domain disparities in Table 3.}
To quantify domain disparities, we calculate the number of high-level skills that must be decomposed into different series of low-level skills for execution in the target domain, compared to the training environment.

For instance, the high-level skill "cook salmon", initially learned in the training environment through a skill-chaining process involving low-level skills such as "grasp salmon" and "place salmon in the microwave", may require adaptation in the target domain. This adaptation might involve a different series of low-level skills "go to refrigerator", and "grasp refrigerator handle", among others, for successful execution. Each instance where a high-level skill's execution strategy deviates from its training environment counterpart contributes to the domain shift quantification. The term `None' refers to scenarios without domain shift, meaning the model is evaluated within the training environment itself. The categories of "Small," "Medium," and "Large" domain shifts correspond to the increasing percentages (20\%, 30\%, and 50\%, respectively) of high-level skills that require adaptation through decomposition into low-level skills in the target domain. 

\section{Trajectory Examples and Prompts}
From Table~\ref{table: trajectory1} to~\ref{table:trajectory3}, we describe the trajectory collected by the VirtualHome agent. Specifically, across each skill step, the inputs and outputs of the task planner and skill critic (e.g., detected object names and their states, retrieved skill semantics and skill candidates, generated skills, inputs and outputs of the critic, and fine-grained instructions) are shown in the Table.  
Following these tables, we also document the prompts used for task planning and skill chaining.

\begin{table*}[p]
\begin{center}
\begin{small}
\adjustbox{max width=\textwidth}{
\begin{tabular}{lc c c c cc}
\toprule
 \multirow{2}{*}{ID} & \multirow{2}{*}{Visual Observation} & \multirow{2}{*}{\begin{tabular}{c} Cross-Domain \\ (Domain-Shift Type) \end{tabular}}  & \multirow{2}{*}{Task} & \multicolumn{2}{c}{Instruction Information} \\
 \cmidrule(rl){5-6} 
 &  &  &  & Instruction Type & Instruction Example \\

\midrule
\multirow{5}{*}{1} & \multirow{5}{*}{\includegraphics[scale=0.11]{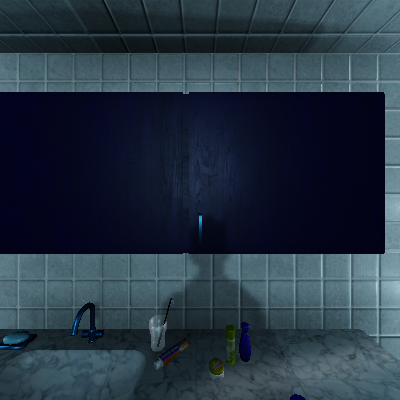}} & \multirow{5}{*}{Object Location (OL)} & \multirow{5}{*}{\begin{tabular}{c} {Put 3 fruits} \\ {on the kitchentable} \end{tabular}} 
& \multirow{5}{*}{Abstract Noun} & \multirow{5}{*}{\begin{tabular}{c} {Grab and put various} \\ {fruits on the kitchentable.} \end{tabular}} \\
  \\
  \\
  \\
  \\
  \\
\multirow{5}{*}{2} & \multirow{5}{*}{\includegraphics[scale=0.11]{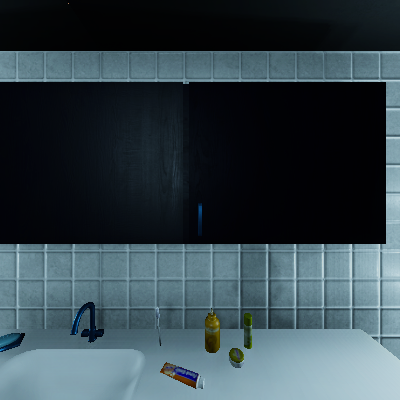}} & \multirow{5}{*}{Room Structure (RS)} & \multirow{5}{*}{\begin{tabular}{c} {Watch tv while} \\ {sitting on sofa} \end{tabular}} 
& \multirow{5}{*}{Abstract Verb} & \multirow{5}{*}{Chill on sofa with tv.} \\
 \\
 \\
 \\
 \\
 \\
\multirow{5}{*}{3} & \multirow{5}{*}{\includegraphics[scale=0.11]{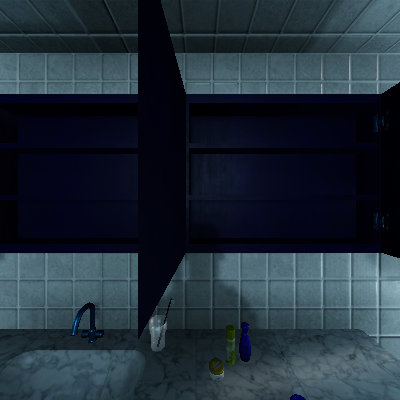}} & \multirow{5}{*}{Physical Attribute (PA)} & \multirow{5}{*}{\begin{tabular}{c} {Put 3 fruits} \\ {in the kitchencabinet} \end{tabular}}
& \multirow{5}{*}{Structured} & \multirow{5}{*}{\begin{tabular}{c} {Grab fruits and put them} \\ {in the kitchencabinet.} \end{tabular}} \\
 \\
 \\
 \\
 \\
 \\
\multirow{5}{*}{4} & \multirow{5}{*}{\includegraphics[scale=0.11]{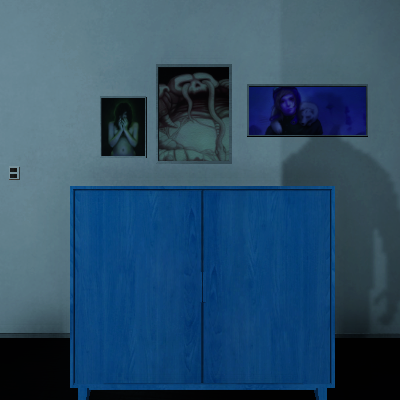}} & \multirow{5}{*}{Object Location (OL)} & \multirow{5}{*}{\begin{tabular}{c} {Put 2 fruits} \\ {in the fridge} \end{tabular}} 
& \multirow{5}{*}{Long Horizon} & \multirow{5}{*}{\begin{tabular}{c} {The kids messed up the kitchen.} \\ {Put food in the fridge.} \end{tabular}} \\
 \\
 \\
 \\
 \\
 \\
\multirow{5}{*}{5} & \multirow{5}{*}{\includegraphics[scale=0.11]{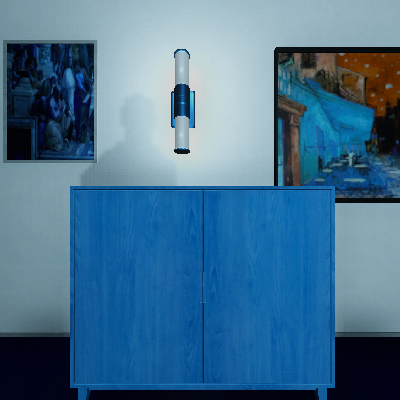}} & \multirow{5}{*}{Room Structure (RS)} & \multirow{5}{*}{\begin{tabular}{c} {Grab book and} \\ {sit on sofa} \end{tabular}}
& \multirow{5}{*}{Abstract Noun} & \multirow{5}{*}{Relax on bunk with book.} \\
 \\
 \\
 \\
 \\
 \\
\multirow{5}{*}{6} & \multirow{5}{*}{\includegraphics[scale=0.11]{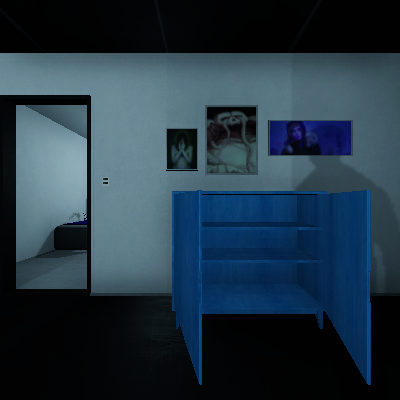}} & \multirow{5}{*}{Physcial Attribute (PA)} & \multirow{5}{*}{\begin{tabular}{c} {Find the food and} \\ {heat it using microwave} \end{tabular}}
& \multirow{5}{*}{Abstract Verb} & \multirow{5}{*}{Heat food.} \\
 \\
 \\
 \\
 \\
 \\
\multirow{5}{*}{7} & \multirow{5}{*}{\includegraphics[scale=0.11]{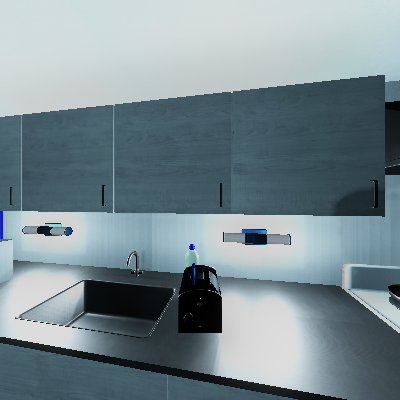}} & \multirow{5}{*}{Object Location (OL)} & \multirow{5}{*}{\begin{tabular}{c} {Put 3 bathroom objects} \\ {in the bathroom cabinet} \end{tabular}}
& \multirow{5}{*}{Structured} & \multirow{5}{*}{\begin{tabular}{c} {Grab bathroom objects} \\ {and put them in cabinet.} \end{tabular}} \\
 \\
 \\
 \\
 \\
 \\
\multirow{5}{*}{8} & \multirow{5}{*}{\includegraphics[scale=0.11]{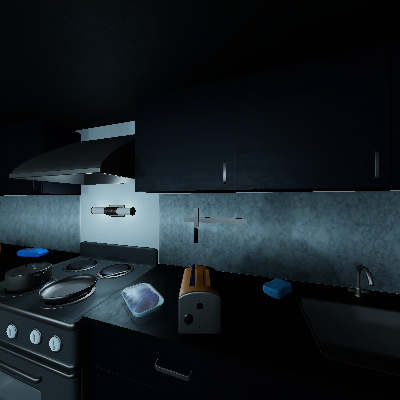}} & \multirow{5}{*}{Room Structure (RS)} & \multirow{5}{*}{\begin{tabular}{c} {Watch tv while} \\ {sitting on sofa} \end{tabular}}
& \multirow{5}{*}{Long Horizon} & \multirow{5}{*}{\begin{tabular}{c} {I finished household chores.} \\ {Relex on sofa and watch tv.} \end{tabular}} \\
 \\
 \\
 \\
 \\
 \\
\multirow{5}{*}{9} & \multirow{5}{*}{\includegraphics[scale=0.11]{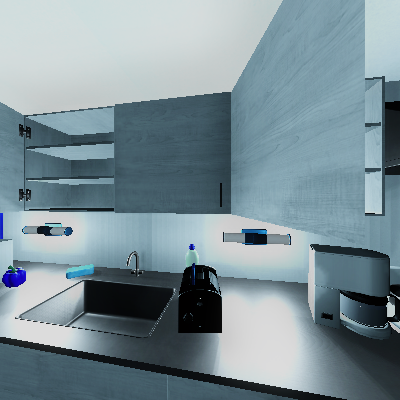}} & \multirow{5}{*}{Physical Attribute (PA)} & \multirow{5}{*}{\begin{tabular}{c} {Put 2 fruits} \\ {in the fridge} \end{tabular}} 
& \multirow{5}{*}{Abstract Noun} & \multirow{5}{*}{Put food in refrigerator.} \\
 \\
 \\
 \\
 \\
 \\
 \bottomrule
\end{tabular}
}
\end{small}
\end{center}

\caption{VirtualHome cross-domain scenario examples}
\label{table: VH scenarios}
\end{table*}

\clearpage

\onecolumn

\begin{table}[h!]
    \begin{tabularx}{\textwidth}{p{4.5cm}X}
    \hline
    \textbf{Skill Step 0} & \\
    Instruction: & I need to clean up the kitchen. Store fruits in the kitchen cabinet. \\
    Visible Object Names: & bananas, kitchen cabinet, fridge, faucet, salmon, peach, kitchen counter drawer, bell pepper, washing sponge, microwave \\
    Visible Objects States: & garbage can is CLOSED, fridge is OPEN, faucet is CLOSED, kitchen cabinet is CLOSED, microwave is CLOSED \\
    Retrieved Skill Semantics: & store fruit in kitchen, put fruit and food in fridge and kitchen cabinet, put away fruit in kitchen cabinet, put away fruit in kitchen cabinet and fridge, put fruit away, fridge food and fruit \\
    Retrieved Skill Candidates: & put peach and plum in fridge, put plum and peach in kitchen cabinet, put fruit in kitchen cabinet, put plum and cupcake in fridge, put apple in kitchen cabinet and peach in fridge \\
    Generated Semantic Skill: & put fruit in kitchen cabinet \\
    Critic Input: & detected objects: bananas, kitchen cabinet, fridge, faucet, salmon, peach, kitchen counter drawer, bell pepper, washing sponge, microwave \\
    & detected states: garbage can is CLOSED, fridge is OPEN, faucet is CLOSED, kitchen cabinet is CLOSED, microwave is CLOSED \\
    & low level skills: put peach and plum in fridge, put plum and peach in kitchen cabinet, put fruit in kitchen cabinet, put plum and cupcake in fridge, put apple in kitchen cabinet and peach in fridge \\
    & current skill: put fruit in kitchen cabinet \\
    Critic Output: & executable: False \\
    & feedback: kitchen cabinet is closed, you need to open the kitchen cabinet \\
    Fine-grained Instruction: & Open kitchen cabinet and put fruit in it \\
    \hline
    \textbf{Skill Step 1} & \\
    Instruction: & Open kitchen cabinet and put fruit in it \\
    Visible Object Names: & bananas, kitchen cabinet, fridge, faucet, salmon, peach, kitchen counter drawer, bell pepper, washing sponge, microwave \\
    Visible Objects States: & garbage can is CLOSED, fridge is OPEN, faucet is CLOSED, kitchen cabinet is CLOSED, microwave is CLOSED \\
    Retrieved Skill Semantics: & put fruit in kitchen cabinet, store bananas and apple in kitchen cabinet, place apple and peach in kitchen cabinet, put cereal and bananas in kitchen cabinet, put bananas in kitchen cabinet and apple on coffee table \\
    Retrieved Skill Candidates: & put peach in kitchen cabinet, open kitchen cabinet and put bananas in kitchen cabinet, open kitchen cabinet and put pie in kitchen cabinet, open kitchen cabinet and put bananas in kitchen cabinet \\
    Generated Semantic Skill: & open kitchen cabinet and put bananas in kitchen cabinet \\
    Critic Input: & detected objects: bananas, kitchen cabinet, fridge, faucet, salmon, peach, kitchen counter drawer, bell pepper, washing sponge, microwave \\
    & detected states: garbage can is CLOSED, fridge is OPEN, faucet is CLOSED, kitchen cabinet is CLOSED, microwave is CLOSED \\
    & low level skills: put peach in kitchen cabinet, open kitchen cabinet and put bananas in kitchen cabinet, open kitchen cabinet and put bananas in kitchen cabinet \\
    & current skill: open kitchen cabinet and put bananas in kitchen cabinet \\
    Critic Output: & executable: True \\
    Executed Skill: & skill: open kitchen cabinet and put bananas in kitchen cabinet \\
    & skill step: walk to bananas, grab bananas, walk to kitchen cabinet, find kitchen cabinet, open kitchen cabinet, put bananas in kitchen cabinet \\
    \hline
    \end{tabularx}
    \caption{Sampled trajectory within the VirtualHome environment.}
    \label{table: trajectory1}
\end{table}

\begin{table}[h!]
    \begin{tabularx}{\textwidth}{p{4.5cm}X}
    \hline
    \textbf{Skill Step 2} & \\
    Instruction: & I need to clean up the kitchen. Store fruits in the kitchen cabinet. \\
    Visible Object Names: & washing sponge, coffee maker, sink, kitchen cabinet, kitchen counter, kitchen counter drawer, toaster, stove, frying pan, coffee pot \\
    Visible Objects States: & coffee maker is CLOSED, kitchen cabinet is CLOSED, kitchen counter is CLOSED, toaster is CLOSED, stove is CLOSED, coffee pot is CLOSED \\
    Retrieved Skill Semantics: & put fruit in kitchen cabinet, put away fruit in kitchen cabinet, store fruit in kitchen, put away fruit in kitchen cabinet and fridge, put fruits in fridge, fridge, put fruit away, put away fruit, store fruit in kitchen cabinet and fridge \\
    Retrieved Skill Candidates: & put apple and peach in kitchen cabinet, place apple in kitchen cabinet, put fruit in kitchen cabinet, put peach and plum in fridge, put plum and peach in kitchen cabinet, put plum and cupcake in fridge \\
    Generated Semantic Skill: & place apple in kitchen cabinet \\
    Critic Input: & detected objects: washing sponge, coffee maker, sink, kitchen cabinet, apple, kitchen counter, kitchen counter drawer, toaster, stove, frying pan, coffee pot \\
    & detected states: coffee maker is CLOSED, kitchen cabinet is OPEN, kitchen counter is CLOSED, toaster is CLOSED, stove is CLOSED, coffee pot is CLOSED \\
    & lower level skills: put apple and peach in kitchen cabinet, place apple in kitchen cabinet, put fruit in kitchen cabinet, put peach and plum in fridge, put plum and peach in kitchen cabinet, put plum and cupcake in fridge \\
    & current skill: place apple in kitchen cabinet \\
    Critic Output: & executable: True \\
    Executed Skill: & skill: place apple in kitchen cabinet \\
    & skill steps: walk to apple, grab apple, walk to kitchen cabinet, put apple in kitchen cabinet, close kitchen cabinet \\
    \hline

    \textbf{Skill Step 3} & \\
    Instruction: & I need to clean up the kitchen. Store fruits in the kitchen cabinet. \\
    Visible Objects: & washing sponge, coffee maker, sink, kitchen cabinet, kitchen counter, kitchen counter drawer, toaster, stove, frying pan, coffee pot, peach \\
    Visible Objects States: & coffee maker is CLOSED, kitchen cabinet is CLOSED, kitchen counter is CLOSED, toaster is CLOSED, stove is CLOSED, coffee pot is CLOSED \\
    Retrieved Skill Semantics: & put fruit in kitchen cabinet, put fruit in fridge and kitchen cabinet, put fruit and food in fridge and kitchen cabinet, put away fruit in kitchen cabinet \\
    Retrieved Skill Candidates: & put apple and peach in kitchen cabinet, put peach and plum in fridge, put plum and peach in kitchen cabinet, store apple in kitchen cabinet and fridge, put apple in kitchen cabinet and peach in fridge \\
    Generated Semantic Skill: & put apple and peach in kitchen cabinet \\
    \hline
    \end{tabularx}
    \caption{Sampled trajectory within the VirtualHome environment.}
    \label{table:trajectory2}
\end{table}

\begin{table}[h!]
    \begin{tabularx}{\textwidth}{p{4.5cm}X}
    \hline
    Critic Input: & detected objects: washing sponge, coffee maker, sink, kitchen cabinet, kitchen counter, kitchen counter drawer, toaster, stove, frying pan, coffee pot, peach \\
              & detected states: coffee maker is CLOSED, kitchen cabinet is CLOSED, kitchen counter is CLOSED, toaster is CLOSED, stove is CLOSED, coffee pot is CLOSED \\ 
              & low level skills: put apple and peach in kitchen cabinet, put peach and plum in fridge, put plum and peach in kitchen cabinet, store apple in kitchen cabinet and fridge, put apple in kitchen cabinet and peach in fridge \\
              & current skill: put apple and peach in kitchen cabinet \\
    Critic Output: & executable: False \\
                   & feedback: kitchen cabinet is closed, you need to open the kitchen cabinet \\
    Fine-grained Instruction: & Open the kitchen cabinet and put apple and peach in the kitchen cabinet \\
    \hline
    \textbf{Skill Step 4} & \\
    Instruction: & I need to clean up the kitchen. Store fruits in the kitchen cabinet. \\
    Visible Objects: & washing sponge, coffee maker, sink, kitchen cabinet, kitchen counter, kitchen counter drawer, toaster, stove, frying pan, coffee pot, peach \\
    Visible Objects States: & coffee maker is CLOSED, kitchen cabinet is CLOSED, kitchen counter is CLOSED, toaster is CLOSED, stove is CLOSED, coffee pot is CLOSED \\
    Retrieved Skill Semantics: & put apple and peach in kitchen cabinet, put fruit in kitchen cabinet, put plum and peach in kitchen cabinet, put peach and plum in kitchen cabinet, put cupcake and peach in kitchen cabinet \\
    Retrieved Skill Candidates: & put apple in kitchen cabinet, open kitchen cabinet and put peach in kitchen cabinet, put peach in kitchen cabinet, put peach in kitchen cabinet, open kitchen cabinet and put pie in kitchen cabinet \\
    Generated Semantic Skill: & open kitchen cabinet and put peach in kitchen cabinet \\
    Critic Input: & detected objects: washing sponge, coffee maker, sink, kitchen cabinet, kitchen counter, kitchen counter drawer, toaster, stove, frying pan, coffee pot, peach \\ 
                & detected states: coffee maker is CLOSED, kitchen cabinet is CLOSED, kitchen counter is CLOSED, toaster is CLOSED, stove is CLOSED, coffee pot is CLOSED \\ 
                & current skill: open kitchen cabinet and put peach in kitchen cabinet \\
    Critic Output: & {executable: True} \\
    Executed Skill: & skill: open kitchen cabinet and put peach in kitchen cabinet \\ 
                    & skill steps: walk to peach, grab peach, walk to kitchen cabinet, find kitchen cabinet, open kitchen cabinet, put peach in kitchen cabinet, close kitchen cabinet \\
    \hline
    \end{tabularx}
    \caption{Sampled trajectory within the VirtualHome environment.}
    \label{table:trajectory3}
\end{table}

\clearpage
\centering
\textbf{Prompt examples utilized by the skill generator in SemGro.}
\begin{lstlisting}[style=mypy]
Examples of common household tasks and their skill sequence to achieve the tasks using skills from Skill Library:

Skill Library: find kitchencabinet, put bananas kitchencabinet, open kitchencabinet, walk kitchencabinet, put bananas kitchentable, find peach, close kitchencabinet, walk bananas, walk cereal, put peach kitchentable, put bananas coffeetable, find coffeetable, find kitchentable, grab bananas
Task: open kitchencabinet and put bananas in kitchencabinet
skill steps: 1. walk bananas 2. grab bananas 3. walk kitchencabinet 4. find kitchencabinet 5. open kitchencabinet 6. put bananas kitchencabinet 7. close kitchencabinet

Skill Library: find bananas, put bananas kitchentable, walk peach, open kitchencabinet, walk coffeetable, put apple kitchencabinet, walk kitchencabinet, close kitchencabinet, walk cereal, find kitchencabinet, put peach kitchentable, put peach kitchencabinet, grab peach
Task: open kitchencabinet and put peach in kitchencabinet
skill steps: 1. walk peach 2. grab peach 3. walk kitchencabinet 4. find kitchencabinet 5. open kitchencabinet 6. put peach kitchencabinet 7. close kitchencabinet

Skill Library: walk peach, walk bananas, walk kitchentable, open kitchencabinet, put apple kitchencabinet, find apple, put peach kitchencabinet, grab peach, walk kitchencabinet, walk apple, find kitchencabinet, find peach, grab apple, grab cereal, close kitchencabinet
Task: put peach in kitchencabinet
skill steps: 1. walk peach 2. find peach 3. grab peach 4. walk kitchencabinet 5. find kitchencabinet 6. open kitchencabinet 7. put peach kitchencabinet 8. close kitchencabinet

Skill Library: put cereal kitchencabinet, find bananas, walk bananas, put bananas kitchentable, walk cereal, walk coffeetable, put bananas kitchencabinet, walk kitchencabinet, find kitchentable, grab peach, find apple, find coffeetable, walk kitchentable, grab bananas, put peach kitchencabinet
Task: find bananas and put bananas on kitchentable
skill steps: 1. walk bananas 2. find bananas 3. grab bananas 4. walk kitchentable 5. find kitchentable 6. put bananas kitchentable

Skill Library: close kitchencabinet, find apple, grab apple, walk apple, walk kitchentable, open kitchencabinet, walk coffeetable, put cereal kitchencabinet, find kitchencabinet, put apple kitchencabinet, walk cereal, find coffeetable, walk bananas, walk kitchencabinet
Task: put apple in kitchencabinet
skill steps: 1. walk apple 2. find apple 3. grab apple 4. walk kitchencabinet 5. find kitchencabinet 6. open kitchencabinet 7. put apple kitchencabinet 8. close kitchencabinet

Skill Library: put apple kitchencabinet, walk cereal, put bananas kitchencabinet, find coffeetable, find kitchentable, walk bananas, grab bananas, open kitchencabinet, put peach kitchentable, put bananas coffeetable, grab peach, walk coffeetable, find peach, put bananas kitchentable, find bananas, walk kitchentable
Task: find bananas and put bananas on coffeetable
skill steps: 1. walk bananas 2. find bananas 3. grab bananas 4. walk coffeetable 5. find coffeetable 6. put bananas coffeetable

Skill Library: find bananas, walk coffeetable, grab cereal, walk bananas, find coffeetable, open kitchencabinet, grab bananas, find kitchencabinet, close kitchencabinet, put bananas coffeetable, walk kitchencabinet, put cereal kitchencabinet, put peach kitchentable, walk cereal
Task: open kitchencabinet and put cereal in kitchencabinet
skill steps: 1. walk cereal 2. grab cereal 3. walk kitchencabinet 4. find kitchencabinet 5. open kitchencabinet 6. put cereal kitchencabinet 7. close kitchencabinet

Choose the next skill from Skill library to successfully accomplish the task.
Skill library: walk bananas, grab bananas, walk kitchencabinet, find kitchencabinet, open kitchencabinet, put bananas kitchencabinet, close kitchencabinet, walk bananas, find bananas, grab bananas, walk kitchencabinet, find kitchencabinet, open kitchencabinet, put bananas kitchencabinet, close kitchencabinet, walk peach, grab peach, walk kitchencabinet, find kitchencabinet, open kitchencabinet, put peach kitchencabinet, close kitchencabinet, walk apple, grab apple, walk kitchencabinet, find kitchencabinet, open kitchencabinet, put apple kitchencabinet, 
skill steps:  1.
\end{lstlisting}
\clearpage
\centering
\textbf{Prompt examples utilized for the skill chaining.}
\begin{lstlisting}[style=mypy]
Examples of common household tasks and their descriptions:
The Task are composed of skill steps chosen from Skill Library,
Skill Library: wash the frying pan, put the candle on the living room shelf,throw away the lime, open window, listen to radio, wash clothes, wash the plate, get my pants and shirt and place on the bed, call a friend, wash mug, wash the rug in washing machine, 
cut apple, make toast, read book under table lamp, brush teeth, open the curtains, put clothes in closet, throw away apple, put all the cutlery in the sink,  put papers in the folder, put my slippers near the bed, take out the trash, microwave chicken, heat cupcakte

Skill Steps: 1. wash the plate 2. wash mug
Task: wash the plate and mug

Skill Steps: 1. brush teeth and wash clothes
Task: brush teeth and wash clothes 

Skill Steps: 1. brush teeth 2. read book under table lamp
Task: read book under table lamp after brushing teeth

Skill Steps: 1. wash the frying pan 2. wash the plate 3. plate mug
Task: wash the kitchenware

Skill Steps: 1. wash clothes 2. wash the rug in washing machine
Task: wash clothes and rug using washing machine

Skill Steps: 1. throw away apple 2. throw away the lime
Task: throw away apple and lime

Skill Steps: 1. cut apple 2. make toast
Task: cut apple and make toast

Skill Steps: 1. make tea and bring it to the coffee table 2. set up coffee table for fruits
Task: set up tea and fruits on coffeetable

Skill Steps 1. put apple in fridge and bring cellphone to sofa 2. put hairproduct in bathroom cabinet and bring paper to sofa
Task: clean up and bring cellphone and paper to sofa

Skill Steps: 1. reheat food 2. prepare chicken for dinner
Task: cook chichen for dinner

Skill Steps: 1. put book back and sit on the sofa 2. watch tv and turn off microwave
Task: organize books and watch tv

Skill Steps: 1. put things back in their place 2. put away mug and lime
Task: clean bedroom 

Skill Steps: 1. reheat cupcake and put it on coffeetable 2. reheat chicken and put it on kitchentable
Task: heat cupcake and chicken and put them on coffeetable

Skill Steps: 1. put away mug and lime 2. eat bread on the sofa and put book back in the bookshelf
Task: clean livingroom

Skill Steps: 1. open the curtains 2. put my slippers near the bed
Task: open the curtains and put my slippers near the bed

Skill Steps: 1. open window 2. call a friend
Task: open window and call a friend

Skill Steps: 1. microwave chicken 2. heat some cupcake
Task: microwave the chicken and heat some cupcake

Skill Steps: 1. microwave chicken 2. heat some cupcake
Task: microwave the chicken and heat some cupcake

Predict the next skill for creating plausible task by correctly choosing from Skill Library:
\end{lstlisting}

\clearpage
\centering
\textbf{Prompt examples utilized for the skill chaining.}
\begin{lstlisting}[style=mypy]
Instructions: give a high-level description for the following steps describing common household tasks.

Skill Steps: 1. wash the plate 2. wash mug
Summary: wash the plate and mug

Skill Steps: 1. brush teeth and wash clothes
Summary: brush teeth and wash clothes 

Skill Steps: 1. brush teeth 2. read book under table lamp
Summary: read book under table lamp after brushing teeth

Skill Steps: 1. wash the frying pan 2. wash the plate 3. plate mug
Summary: wash the kitchenware

Skill Steps: 1. wash clothes 2. wash the rug in washing machine
Summary: wash clothes and rug using washing machine

Skill Steps: 1. make tea and bring it to the coffee table 2. set up coffee table for fruits
Summary: set up tea and fruits on coffeetable

Skill Steps 1. put apple in fridge and bring cellphone to sofa 2. put hairproduct in bathroom cabinet and bring paper to sofa
Summary: clean up and bring cellphone and paper to sofa

Skill Steps: 1. reheat food 2. prepare chicken for dinner
Summary: cook chichen for dinner

Skill Steps: 1. put book back and sit on the sofa 2. watch tv and turn off microwave
Summary: organize books and watch tv

Skill Steps: 1. put things back in their place 2. put away mug and lime
Summary: clean bedroom 

Skill Steps: 1. reheat cupcake and put it on coffeetable 2. reheat chicken and put it on kitchentable
Summary: heat cupcake and chicken and put them on coffeetable

Skill Steps: 1. put away mug and lime 2. eat bread on the sofa and put book back in the bookshelf
Summary: clean livingroom

Skill Steps: 1. throw away apple 2. throw away the lime
Summary: throw away apple and lime

Skill Steps: 1. cut apple 2. make toast
Summary: cut apple and make toast

Skill Steps: 1. open the curtains 2. put my slippers near the bed
Summary: open the curtains and put my slippers near the bed

Skill Steps: 1. open window 2. call a friend
Summary: open window and call a friend

Skill Steps: 1. microwave chicken 2. put chicken on coffeetable
Summary: heat chicken and put it on coffeetable


Summarize the skill steps and give a plausible name for the household task executed.
\end{lstlisting}




\end{document}